%% file: main.tex
\pdfoutput=1
\documentclass[11pt,x11names]{article}

\usepackage{acl}

\usepackage{xcolor}

\usepackage{times}
\usepackage{latexsym}

\usepackage{bold-extra}

\usepackage[T1]{fontenc}
\usepackage[utf8]{inputenc}

\usepackage{microtype}

\usepackage{inconsolata}

\usepackage{graphicx}

\usepackage{mdframed}
\mdfdefinestyle{mdframedlisting}{
 backgroundcolor=black!10,
 linecolor=white,
 innertopmargin=-2mm,
 innerbottommargin=0mm,
 innerrightmargin=0mm,
 innerleftmargin=0mm,
}
\usepackage{listings}
\usepackage{colortbl} % for \cellcolor
\usepackage{todonotes}

\usepackage{marginnote}

\usepackage{xspace}

\usepackage{amsmath}
\usepackage{cleveref}
\usepackage{arydshln}
\usepackage{booktabs}
\usepackage{makecell}
\usepackage{multicol}
\usepackage{MnSymbol} % for \bigstar
\usepackage{multirow}
\usepackage{enumitem}
\setlist[itemize]{noitemsep,left=0ex,topsep=0ex}

\usepackage{tikz}
\usepackage{pgfplots}\pgfplotsset{compat=1.18}

\newcommand{\XtoX}[2]{\makebox[3.5mm][c]{#1}$\rightarrow$\makebox[3.5mm][c]{#2}\xspace}
\newcommand{\EnZh}{\XtoX{En}{Zh}}
\newcommand{\EnDe}{\XtoX{En}{De}}
\newcommand{\ZhEn}{\XtoX{Zh}{En}}
\newcommand{\DeEn}{\XtoX{De}{En}}

\newcommand{\problemtypetechnical}{\raisebox{0.3mm}{\small [technical]}}
\newcommand{\problemtypedata}{\raisebox{0.3mm}{\small [data]}}
\newcommand{\problemtypeusage}{\raisebox{0.3mm}{\small [usage]}}

\usepackage{soul}
\newcommand{\hlc}[2][yellow]{{%
    \colorlet{foo}{#1}%
    \sethlcolor{foo}\hl{#2}}%
}

\title{Pitfalls and Outlooks in Using COMET}
\makeatletter\def\Hy@Warning#1{}\makeatother
\let\svthefootnote\thefootnote
\newcommand\blankfootnote[1]{%
  \let\thefootnote\relax\footnotetext{#1}%
  \let\thefootnote\svthefootnote%
}

\newcommand\scalemath[2]{\scalebox{#1}{\mbox{\ensuremath{\displaystyle #2}}}}
\newcommand{\midstar}{\scalemath{0.5}{\bigstar}}

\author{
    Vilém Zouhar$^{\midstar}$\,$^{1}$\quad
    Pinzhen Chen$^{\midstar}$\,$^{2}$\quad
    Tsz Kin Lam$^2$\quad
    Nikita Moghe$^2$\quad
    Barry Haddow$^2$
    \\[1em]
    $^1$ETH Zurich \qquad   
    $^2$University of Edinburgh \\
    \texttt{\href{mailto:vzouhar@ethz.ch}{vzouhar@ethz.ch}} \quad \texttt{\{\href{mailto:pinzhen.chen@ed.ac.uk}{pinzhen.chen},\href{mailto:tlam@ed.ac.uk}{tlam},\href{mailto:nikita.moghe@ed.ac.uk}{nikita.moghe},\href{mailto:bhaddow@ed.ac.uk}{bhaddow}\}@ed.ac.uk}
}

\begin{document}
\maketitle
\blankfootnote{$^{\midstar}$Equal contributions.}
\blankfootnote{$^0$Code: \href{https://github.com/PinzhenChen/sacreCOMET}{github.com/PinzhenChen/sacreCOMET}}

\begin{abstract}
% Since its introduction, t
The COMET metric has blazed a trail in the machine translation community given its strong correlation with human judgements of translation quality.
Its success stems from being a pre-trained multilingual model finetuned for quality assessment.
However, it being a neural metric also gives rise to a set of pitfalls that may not be widely known. 
We investigate these unexpected behaviours from three aspects:
1) technical: obsolete software versions and compute precision;
2) data: empty content, language mismatch, and translationese at test time as well as distribution and domain biases in training;
3) usage and reporting: multi-reference support and model referencing in the literature. 
All of these problems imply that COMET scores may be not incomparable between papers or technical setups and we put forward our perspective on fixing each issue.
Furthermore, we release the \texttt{sacreCOMET} package that can generate a signature for the software and model configuration as well as an appropriate citation.
The goal of this work is to help the community make more sound use of the COMET metric.
\end{abstract}

\section{Introduction}

Automated metrics provide a cheap and scalable way of evaluating and benchmarking NLP models.
In machine translation (MT), the evaluation protocol has moved from string matching metrics \citep[BLEU, TER, chrF, inter alia;][]{papineni-etal-2002-bleu, snover-etal-2006-study, popovic-2015-chrf} to trained neural metrics \citep{shimanaka-etal-2018-ruse,takahashi-etal-2020-automatic, rei-etal-2020-comet, sellam-etal-2020-bleurt} with COMET being widely adopted.
% the most widely used option in the community.
The trained metrics have been shown to correlate much better with human judgement \citep{freitag-etal-2021-results, freitag-etal-2022-results, freitag-etal-2023-results}, making them more reliable in estimating translation quality and ranking translation systems.

Nonetheless, the solution to translation evaluation is yet to be perfected.
One problem is the haphazard use of the metric. 
Previously, \citet{post-2018-call} showed that different usages and implementations of BLEU, e.g. tokenization and smoothing, lead to inconsistencies in scores.
% We suspect that sensitivities can arise from several stages of using neural metrics too---e.g.\ obtaining COMET involves running a particular checkpoint with the COMET software on some particular hardware.
We suspect that the use of COMET might be sensitive to misconfigurations too, resulting in unexpected %or nonreproducible % commented out because I think the "incorrect behaviour" can still be reproducible if the same config is passed.
behaviours.
Furthermore, trained MT metrics are optimized on a limited amount of data (usually valid machine translations), leading to overfitting and reduced robustness against corner cases.
%often fit on limited data intended to be valid translations, leading to reduced robustness owing to its nature as an optimized machine learning model. 
% list nine possible problems with using COMET, provide ways to address some of them, including a helpful package \texttt{SacreCOMET}, and suggest further research options.
% \medskip
% \noindent
Contributions of this work are listed as follows:
\begin{itemize}
\item we reveal nine problems spanning technical issues, data biases, and model reporting;
\item we show that inconsistent use of COMET leads to non-comparable scores across papers or setups;
\item we release the \href{https://pypi.org/project/sacrecomet/}{\texttt{sacreCOMET}} package for better reporting and reproducibility;
\item we provide directions for future work on building learned metrics.
\end{itemize}

\section{Background and Setup}

\vspace{-2mm}

\paragraph{Metric background.}

Publicly available human judgements of translation quality come from shared task annotation campaigns, where translations are evaluated with some annotation protocol. 
From 2017, in WMT, the protocol was a variant of direct assessment \citep[DA;][]{graham-etal-2013-continuous} which has annotators providing a number from 0 (lowest) to 100 (highest) as the segment quality.
This has been subsequently replaced by MQM and ESA protocols \citep{lommel-etal-2014-using, kocmi2024errorspanannotationbalanced}, though DA remains the most abundant data source for neural metric training.

Automated metrics aim to yield scores that correlate with human judgements of translations.
Most metric scores are computed at the segment level and then aggregated at the system level to e.g. obtain system comparison.
The evaluation of metrics is done with respect to the human judgements.
% either at the segment level or system level
% An example of a metric is character-level n-gram f-score (chrF, \citealp{popovic-2015-chrf}), which computes character n-gram matches between the hypothesis and the human reference.
% \Vilem{Here explain DA/MQM a little bit?}

\paragraph{COMET models.}
Metrics such as chrF or BLEU are heuristic algorithms that match n-grams between the translation and the reference to compute a score.
In contrast, COMET is a machine learning model fine-tuned from a pre-trained multilingual language model, e.g. XLM-R \citep{conneau-etal-2020-unsupervised}, with an additional regression head.
A reference-based COMET model learns to regress from a tuple of [source, hypothesis, reference] to the human judgement score (from previous evaluation campaigns) at the segment level.
The quality estimation (reference-free) version of COMET is prepared by omitting the reference from the input.

Most issues in this work are demonstrated using two COMET checkpoints unless noted otherwise: the reference-based \href{https://huggingface.co/Unbabel/wmt22-comet-da}{COMET$^\mathrm{DA}_\mathrm{22}$} and the reference-free \href{https://huggingface.co/Unbabel/wmt22-cometkiwi-da}{COMET$^\mathrm{kiwiDA}_\mathrm{22}$} \citep{rei-etal-2022-comet}.
% \footnote{\href{}{huggingface.co/Unbabel/wmt22-cometkiwi-da}} 
Both metrics output to a normalized range between~0~and~1. 
The COMET framework \texttt{unbabel-comet} is of version 2.2.2 except when we test how different software versions affect COMET scores.
% VZ: Afaik there's no difference in this
% Scores are obtained via \texttt{comet-score}.

\paragraph{Data setup.}
We base our experiments on the general domain translation and metrics shared tasks of WMT from 2023 \citep{kocmi-etal-2023-findings,freitag-etal-2023-results}.
The translation directions in the paper are centred around En$\leftrightarrow$De and En$\leftrightarrow$Zh, though we occasionally include other translation directions for demonstrative purposes.
% Replication on other translation directions can be found in the appendix. % Patrick: think this is not very important

Whenever possible, we divide all scores, including DA and model outputs, such that their output is between 0 and 1.

\section{Problems}

In this section, we identify and test nine possible pitfalls or curious behaviours with COMET which are not all well-studied.
In three groups, these are:
\begin{itemize}
\item \textbf{Technicality}: obsolete Python and COMET software versions as well as compute precisions could lead to inaccurate score computation.
\item \textbf{Training and test data}: COMET as a neural metric, might be derailed by empty hypotheses, language mismatch, and translationese at test time. It may also follow the training data biases.
\item \textbf{Tool usage and score interpretation}: 
COMET has no defined way of equipping multiple references when available which leaves room for research.
From a bibliometric perspective, we reveal that some literature omits a clear reference to the checkpoint version or citation.
\end{itemize}
In addition, we discuss some final issues that need more attention from the community.

\begin{table}[t]
 \centering\small
 \begin{tabular}{cccc}
 \toprule
 \multicolumn{1}{l}{Python} & 3.7.16 & 3.8.11 & 3.12.4 \\
 \multicolumn{1}{l}{unbabel-comet} & 1.1.2 & 2.2.2 & 2.2.2 \\
 \midrule
 \EnDe & \textcolor{Firebrick3!85}{0.796} & \textcolor{RoyalBlue4!85}{0.837} & \textcolor{RoyalBlue4!85}{0.837} \\
 \EnZh & \textcolor{Firebrick3!85}{0.911} & \textcolor{RoyalBlue4!85}{0.862} & \textcolor{RoyalBlue4!85}{0.862} \\
 \DeEn & \textcolor{Firebrick3!85}{0.851} & \textcolor{RoyalBlue4!85}{0.855} & \textcolor{RoyalBlue4!85}{0.855} \\
 \ZhEn & \textcolor{Firebrick3!85}{0.795} & \textcolor{RoyalBlue4!85}{0.803} & \textcolor{RoyalBlue4!85}{0.803} \\
 \bottomrule
 \end{tabular}
 \vspace{-1ex}
 \caption{COMET$^\mathrm{DA}_\mathrm{22}$ scores for WMT 23 Online-A under different package versions.}
 \vspace{-2ex}
 \label{tab:versions}
\end{table}

\subsection{Software versions \problemtypetechnical}
\label{sec:software_version}

The official installation of the COMET package requires Python 3.8 or above.\footnote{\href{https://github.com/Unbabel/COMET}{github.com/Unbabel/COMET} \texttt{332dfb0} as of Aug 2024.}
We demonstrate that neglecting this would lead to unexpected scores because the same COMET checkpoint can produce vastly different scores with previous COMET framework versions that are no longer supported.

Under several Python versions, executing the following code leads to different COMET package (\texttt{unbabel-comet}) versions being installed.
Running the framework for translation evaluation will subsequently result in false conclusions as shown in \Cref{tab:versions}'s evaluation on WMT23 tests.
The direct cause is that Python 3.7, which has been discontinued, only supports \texttt{unbabel-comet} versions up to 1.1.2. Nonetheless, we caution that the underlying factor is the version of \texttt{unbabel-comet} rather than Python.

\begin{mdframed}[style=mdframedlisting]
\begin{lstlisting}
$ pip install pip --upgrade
$ pip install unbabel-comet --upgrade

# will install
# unbabel-comet==1.1.2 under Python 3.7.16
# unbabel-comet==2.2.2 under Python 3.8.11
# unbabel-comet==2.2.2 under Python 3.12.4
\end{lstlisting}
\end{mdframed}

\paragraph{Recommendation.}
Updating both Python and \texttt{unbabel-comet} to their latest versions is helpful and reporting the toolkit version aids reproducibility.
% The package version should also be reported.

\begin{table}[t]
\centering\small\setlength{\tabcolsep}{0.5ex}
% \resizebox is needed otherwise it's still too wide
\resizebox{\linewidth}{!}{
\begin{tabular}{c@{\hskip -1ex}c@{\hskip 0.25ex}c@{\hskip 0.5ex}c@{\hskip 1.5ex}c@{\hskip 1.5ex}c@{\hskip 1ex}c@{\hskip 1ex}c}
\toprule
   & \multicolumn{1}{c}{Precision} & COMET$_\mathrm{22}^\mathrm{DA}$ & MAE & ${\tau}_{c}$ & Acc & Time (s) \\
\midrule
\multirow{5}{*}{\EnDe} & \multirow{2}{*}{GPU} FP32 & 0.822 & 10.4  & 0.274 & 0.885  & \phantom{0}113  \\
   & \phantom{GPU} FP16 & 0.822 & 10.4  & 0.274 & 0.885  & \phantom{00}55 \\
\cmidrule(l){2-7}
   & \multirow{3}{*}{CPU} FP32 & 0.822 & 10.4  & 0.274 & 0.885  & 2262 \\
   & \phantom{CPU} FP16 & 0.822 & 10.4  & 0.274 & 0.885  & 2403 \\
   & \phantom{CPU} QINT8   & 0.852 & 10.8 & 0.109 & 0.385  & 1856 \\
\midrule
\multirow{5}{*}{\DeEn} &  \multirow{2}{*}{GPU} FP32 & 0.841 & 9.98  & 0.296 & 0.901  & \phantom{00}87   \\
   & \phantom{GPU} FP16 & 0.841 & 9.98  & 0.296 & 0.901  & \phantom{00}48 \\
\cmidrule(l){2-7}
   & \multirow{3}{*}{CPU} FP32 & 0.841 & 9.98  & 0.296 & 0.901  & 1674 \\
   & \phantom{CPU} FP16 & 0.841 & 9.99  & 0.296 & 0.901  & 1758 \\
   & \phantom{CPU} QINT8   & 0.860  & 10.7  & 0.164 & 0.516  & 1249 \\
\midrule
\multirow{5}{*}{\EnZh} &  \multirow{2}{*}{GPU} FP32 & 0.842 & 11.7  & 0.290  & 0.933  & \phantom{0}111  \\
   & \phantom{GPU} FP16 & 0.842 & 11.7  & 0.290  & 0.933  & \phantom{0}102  \\
\cmidrule(l){2-7}
   & \multirow{3}{*}{CPU} FP32 & 0.842 & 11.7  & 0.290  & 0.933  & 1751 \\
   & \phantom{CPU} FP16 & 0.842 & 11.7  & 0.290  & 0.933  & 1710 \\
   & \phantom{CPU} QINT8   & 0.881 & 13.2  & 0.031   & 0.608  & 1258 \\
\midrule
\multirow{5}{*}{\ZhEn} &  \multirow{2}{*}{GPU} FP32 & 0.799 & 9.95  & 0.153 & 0.717  & \phantom{0}113  \\
   & \phantom{GPU} FP16 & 0.799 & 9.95  & 0.153 & 0.725  & \phantom{00}86 \\
\cmidrule(l){2-7}
   & \multirow{3}{*}{CPU} FP32 & 0.799 & 9.95  & 0.153 & 0.717  & 1936 \\
   & \phantom{CPU} FP16 & 0.799 & 9.95  & 0.153 & 0.717  & 1995 \\
   & \phantom{CPU} QINT8   & 0.872 & 10.5  & 0.081   & 0.475  & 1351 \\
\bottomrule
\end{tabular}
}
\vspace{-2ex}
\caption{System ranking with quantization on GPU and CPU. COMET$_\mathrm{22}^\mathrm{DA}$ is the absolute model score; MAE, $\tau_c$, and Acc are mean average error, correlation, and accuracy with respect to human judgements; Time refers to computation time in seconds.}
\vspace{-1ex}
\label{tab:quantization}
\end{table}

\subsection{Numerical precision \problemtypetechnical}
\label{sec:compute_precision}

Model quantization represents a model using lower-numerical precision data types so that the model consumes less memory and model passes can be computed faster. % it can be faster due to both a larger usable batch size and a lower precision.
Such improvement in inference is directly beneficial to deployment efficiency; it is also useful in other complex procedures involving COMET scoring, such as data filtering, re-ranking, and Minimum Bayes Risk (MBR) decoding \citep{kumar-byrne-2004-minimum}.

Despite the aforementioned advantages, model quantization is not a feature supported by the current COMET framework by default.
We thus make minimal modifications to the software and investigate the effect of numerical precision on COMET scores on both CPU (FP32, FP16, and QINT8) and GPU (FP32 and FP16).
When using FP16, we first load the model weights to FP32, followed by \texttt{.half()} call.
% After that, we call \texttt{.half()}~to explicitly convert the weights of the encoder, i.e., \texttt{XLMRobertaModel()} from \texttt{transformers}.
This is because loading the weights directly in FP16 still incorrectly results in FP32 precision.
For CPU inference with dynamic QINT8, we apply the quantization module \texttt{torch.ao.quantization} \href{https://pytorch.org/docs/stable/quantization.html}{from PyTorch}.
% before \texttt{model.predict()}.

\begin{table*}[t]
 \centering\small
 % \resizebox{\linewidth}{!}{
 \setlength{\tabcolsep}{1ex}
 \begin{tabular}{lcccccccc}
 \toprule
 & \multicolumn{4}{c}{COMET$^\mathrm{DA}_\mathrm{22}$} & \multicolumn{4}{c}{\hspace{5mm}COMET$^\mathrm{kiwiDA}_\mathrm{22}$} \\
\cmidrule(lr){2-5}\cmidrule(lr){6-9}
Hypothesis ($\downarrow$) & \EnDe & \EnZh & \DeEn & \ZhEn & \EnDe & \EnZh & \DeEn & \ZhEn \\
\midrule
Real system (Online-A) & 0.837 & 0.862 & 0.855 & 0.803 & 0.800 & 0.791 & 0.794 & 0.787 \\
Empty hypothesis & 0.335 & 0.392 & 0.353 & 0.374 & 0.315 & 0.319 & 0.537 & 0.467 \\
Random hypothesis (fluent) & 0.373 & 0.434 & 0.334 & 0.350 & 0.333 & 0.341 & 0.447 & 0.391 \\
Random hypothesis (shuffled words) & 0.244 & 0.419 & 0.264 & 0.347 & 0.232 & 0.325 & 0.307 & 0.385 \\
\bottomrule
\end{tabular}
% }
\vspace{-1ex}
\caption{Absolute average COMET scores for WMT23 Online-A, empty hypotheses, and random sentences. Random sentences are either fluent but irrelevant or perturbed with words shuffled and thus non-fluent.}\vspace{-1ex}
\label{tab:empty}
\end{table*}

We use AMD Ryzen 9 5900X with NVIDIA GeForce RTX 3090 for GPU inference and a batch size of 8 in all settings (in practice a quantized model makes room for a larger batch size).
\Cref{tab:quantization} summarises the effect of numerical precision.
In addition to reporting COMET scores, we also report (1) inference time in seconds (sec) as an efficiency measure; and (2) segment-level mean absolute error (MAE), segment-level Kendall's tau-c ($\tau_{c}$), and system-level pairwise accuracy (Acc).
Everything is compared to the human DA scores either on segment- or system-level.
Technically, Kendall's $\tau_{c}$ calculates rank correlation on an ordinal scale with adjustments for ties and pairwise accuracy computes the proportion of system pairs that have the same ordering by a metric as by humans.

Our results show that there is no meaningful difference between FP32 and FP16 in both CPU and GPU devices up to 3 significant figures.
On GPU, FP16 is about 30\% faster in time, but unsurprisingly it does not provide any speed-up on CPU.
Interestingly, on the CPU, dynamic QINT8 gives systematically higher COMET scores and shorter running times than FP32 and FP16. 
However, the much lower ${\tau}_{c}$ and pairwise accuracy indicate the lack of reliability at this precision. 
In addition to precision, we explored the effect of batch size and the choice of GPU or CPU during inference with results listed \Cref{sec:gpucpu_batch_size}.
Whilst there are some fluctuations, they are mostly negligible.
However, lower precision allows for higher batch size which usually directly corresponds to speed-up.

\paragraph{Recommendation.} If GPU is available, it is feasible to run COMET with FP16 with a larger batch size for much faster inference without any quality loss.
Otherwise, FP32 should be used.

\subsection{Empty hypothesis \problemtypedata}
\label{sec:empty_hypothesis}

An empty translation (a string of length 0) gets penalized heavily by string-based metrics because an empty string has zero surface overlap with the reference.
However, neural metrics provide no such guarantee.
We show that COMET assigns a positive instance-level score even if the hypothesis is an empty string as corroborated by \citet{lo-etal-2023-metric}.
Problematically, this score can even occasionally be higher than that of a genuine system translation.

In \Cref{tab:empty}, we list COMET scores for system Online-A's hypotheses at WMT23 and a file full of empty lines. 
Furthermore, we compare them with completely incorrect translations to explicate the score magnitude in two ways:
\begin{itemize}
\item \textit{Random hypothesis:} we shuffle WMT22's reference files at the sentence level in the respective translation directions.
This provides us with high-quality human-written sentences.
We sub-sample or over-sample if the number of lines in WMT22 is larger or smaller than the WMT23 size.
\item \textit{Random hypothesis (shuffled words):} we further shuffle the words at each line in the sentence-shuffled files, generating nonsensical sentences.
\end{itemize}
\noindent
Sentence-shuffled hypotheses can be seen as fluent but extremely inadequate sentences whereas word-shuffled sentences are neither fluent nor adequate.

We observe that sentence-shuffled hypotheses attain comparable scores to empty ones, but word-shuffled hypotheses have the lowest scores across the majority of the translation directions.
Empty and shuffled hypotheses, despite having much lower COMET than the valid translations, would not be assigned zero scores by COMET$^\mathrm{DA}_\mathrm{22}$ or COMET$^\mathrm{kiwiDA}_\mathrm{22}$, showing that COMET is more lenient than string overlap-based metrics in penalizing such irregularities.

\begin{figure}[t]
\centering\small
\input{img/plot-zh-en-wmt22da.tex}\hfill%
\input{img/plot-zh-en-wmt22da-kiwi.tex}\\
\input{img/distributions_custom_legend.tex}
\vspace{-1ex}
\caption{Distribution of instance-level scores for empty and baseline translations (x-axis: score; y-axis: count). See other translation directions in \Cref{app:empty-distribution}.}
\label{fig:empty-distribution-small}
\end{figure}
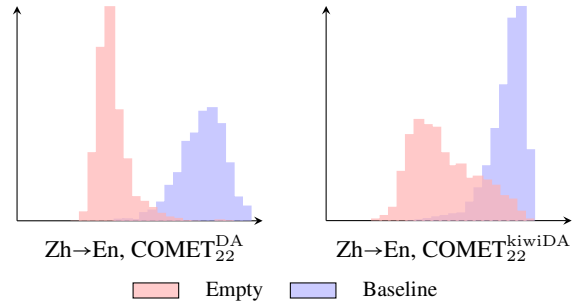

We count the number of empty lines that score better than a translation from Online-A in \Cref{tab:empty-count}. 
We observe roughly 0.25\% such cases for most translation directions except for De$\rightarrow$En's COMET$^\mathrm{kiwiDA}_\mathrm{22}$ score situating at 1.45\%. 
Further, in \Cref{fig:empty-distribution-small} we plot the distributions of COMET scores for Zh$\rightarrow$En's empty and genuine translations with other translation directions in \Cref{app:empty-distribution}. 
Noticeable overlaps are observed for COMET$^\mathrm{kiwiDA}_\mathrm{22}$ when translating into English.

Adhering to the DA protocol guidelines, this is not the proper behaviour because an empty hypothesis should receive a score of 0, to match \textit{no meaning preserved}.
This however is unsurprising with COMET, which has likely not seen empty hypotheses during training that would have received a score of 0 from a human annotator.
%From the perspective of COMET, it is not surprising because during training it has likely not seen empty hypotheses, which would have been assigned 0 by a human annotator. 
Finally, even by relaxing the 0-score expectation, the metric should still assign the same score to all empty hypotheses regardless of the source. Since the distributions of empty hypotheses are nowhere close to a single vertical bar in \Cref{fig:empty-distribution-small}, it exposes the issue that segment-level COMET scores oddly hinge on the source sentence, as noted by \citet{sun-etal-2020-estimating}.

\begin{table}[t]
\centering\small
\begin{tabular}{ccc}
\toprule
 & \multicolumn{2}{c}{translation < empty} \\
\cmidrule(lr){2-3}
 & COMET$^\mathrm{DA}_\mathrm{22}$ & COMET$^\mathrm{kiwiDA}_\mathrm{22}$ \\
\midrule
\EnDe & 0 / 558\phantom{0} & 2 / 558\phantom{0} \\
\XtoX{En}{Ru} & 1 / 2075 & 0 / 2075 \\
\XtoX{En}{Uk} & 2 / 2075 & 1 / 2075 \\
\EnZh & 6 / 2075 & 2 / 2075 \\
\DeEn & 1 / 550\phantom{0} & 8 / 550\phantom{0} \\
\XtoX{Ru}{En} & 5 / 1724 & 6 / 1724 \\
\XtoX{Uk}{En} & 1 / 1827 & 5 / 1827 \\
\ZhEn & 5 / 1977 & 1 / 1977 \\
\bottomrule
% \EnDe & 0.000\% & 0.358\% \\
% \XtoX{En}{Ru} & 0.048\% & 0.000\% \\
% \XtoX{En}{Uk} & 0.096\% & 0.048\% \\
% \EnZh & 0.289\% & 0.096\% \\
% \DeEn & 0.182\% & 1.455\% \\
% \XtoX{Ru}{En} & 0.290\% & 0.348\% \\
% \XtoX{Uk}{En} & 0.055\% & 0.274\% \\
% \ZhEn & 0.253\% & 0.051\% \\#
\end{tabular}\vspace{-1ex}
\caption{Proportion of WMT23 Online-A's translations that are worse than an empty line for the same source, displayed as ``empty/total''.}
\vspace{-1ex}
\label{tab:empty-count}
\end{table}

\paragraph{Recommendation.}
Force empty hypotheses to have 0 scores before aggregating.
Also, a string-based metric like BLEU or chrF should be used to catch similarly malformed hypotheses.

\subsection{Hypothesis language mismatch \problemtypedata}
String overlap-based metrics can also score a hypothesis in a language different from the reference almost zero, especially with script mismatch.
However, even for the reference-based COMET, there is no explicit way to enforce the intended target language.
% given it is an embedding-based metric.
% In the field of translation and translation evaluation, 
This poses an increasingly pronounced problem, especially for multilingual translation models as well as the recent large language models, in which the generated language cannot be as easily controlled \citep{10.5555/3618408.3620130}.
% as it could be in encoder-decoder models 

We conduct experiments to understand if translation outputs in an incorrect language impact the score, and whether different mismatching languages can lead to distinct patterns.
We use the translation directions En$\rightarrow$Ru, En$\rightarrow$Uk, and En$\rightarrow$Zh in WMT23 which share the same English source input.
Having Online-A's output in all three directions, we substitute hypotheses in a particular translation direction with those from another direction.
A similar hypothesis was presented by \citet{amrhein-etal-2022-aces} which suggested that COMET metrics are not robust to hypothesis language mismatch. Our experiment setup offers a more detailed evaluation setup than their contrastive setup.

\Cref{tab:mismatch} presents COMET$^\mathrm{DA}_\mathrm{22}$ scores for translations in correct and incorrect languages, as well as empty lines and random sentences in the correct language as ``baselines'', deemed as completely wrong translations. The pattern shows that when the hypothesis is in a language distant from the reference, the COMET score declines much more than when the hypothesis is in a similar language.
A more concerning issue is that even when the hypotheses' language is completely wrong, the resulting COMET score can still be vastly higher than empty hypotheses or random sentences in the correct language.
We omit COMET$^\mathrm{kiwiDA}_\mathrm{22}$ because it does not have a mechanism to read a reference (language) making it inherently incapable of distinguishing output languages. 

\paragraph{Recommendation.}
Run language identification and set hypotheses in an unexpected language to have a 0 COMET score before aggregating them system-level.
Also, check with a string overlap-based metric like BLEU or chrF.

\begin{table}[t]
\centering\small
 % \setlength{\tabcolsep}{0.75ex}
 % \resizebox{\linewidth}{!}{
 \begin{tabular}{lcccccc}%cccccc}
 \toprule
  % & \multicolumn{6}{c}{COMET$^\mathrm{DA}_\mathrm{22}$} \\%& \multicolumn{6}{c}{COMET$^\mathrm{kiwiDA}_\mathrm{22}$} \\
 % \cmidrule(lr){2-7}%\cmidrule(lr){8-13}
 % \multirow[b]{2}{*}{\makecell{\null\\Hyp. lang. ($\rightarrow$)}}
 % \multirow[b]{2}{*}{\makecell{Mismatch ($\downarrow$)\\Hyp. lang. ($\rightarrow$)}}
 & \multicolumn{2}{c}{\XtoX{En}{Ru}} & \multicolumn{2}{c}{\XtoX{En}{Uk}} & \multicolumn{2}{c}{\XtoX{En}{Zh}} \\%& \multicolumn{2}{c}{en-ru} & \multicolumn{2}{c}{en-uk} & \multicolumn{2}{c}{en-zh} \\
 % \cmidrule(lr){2-3}\cmidrule(lr){4-5}\cmidrule(lr){6-7}%\cmidrule(lr){8-9}\cmidrule(lr){10-11}\cmidrule(lr){12-13}
  % & Uk & Zh & Ru & Zh & Ru & Uk \\%& uk & zh & ru & zh & ru & uk \\
 \midrule
Correct lang. & \multicolumn{2}{c}{0.853} & \multicolumn{2}{c}{0.832} & \multicolumn{2}{c}{0.862} \\
\cmidrule(lr){1-7}
\multirow{2}{*}{\makecell{Incorrect\\target lang.}} & \multicolumn{2}{c}{Uk: 0.797} & \multicolumn{2}{c}{Ru: 0.807} & \multicolumn{2}{c}{Ru: 0.655} \\
%Incorrect lang.
& \multicolumn{2}{c}{Zh: 0.536} & \multicolumn{2}{c}{Zh: 0.540} & \multicolumn{2}{c}{Uk: 0.644} \\
\cmidrule(lr){1-7}
%& 0.814 & 0.814 & 0.799 & 0.799 & 0.791 & 0.791 \\
% \phantom{10}1\% & 0.853 & 0.850 & 0.832 & 0.830 & 0.860 & 0.860 \\%& 0.814 & 0.814 & 0.799 & 0.799 & 0.791 & 0.791 \\
% \phantom{10}2\% & 0.852 & 0.847 & 0.832 & 0.827 & 0.858 & 0.858 \\%& 0.813 & 0.813 & 0.799 & 0.799 & 0.791 & 0.791 \\
% \phantom{10}4\% & 0.852 & 0.841 & 0.831 & 0.821 & 0.854 & 0.854 \\%& 0.813 & 0.813 & 0.800 & 0.799 & 0.792 & 0.791 \\
% \phantom{10}8\% & 0.849 & 0.829 & 0.830 & 0.810 & 0.846 & 0.845 \\%& 0.812 & 0.812 & 0.800 & 0.799 & 0.792 & 0.791 \\
% \phantom{1}16\% & 0.845 & 0.804 & 0.828 & 0.788 & 0.829 & 0.828 \\%& 0.812 & 0.810 & 0.801 & 0.798 & 0.794 & 0.792 \\
% \phantom{1}32\% & 0.836 & 0.753 & 0.825 & 0.741 & 0.797 & 0.794 \\%& 0.809 & 0.806 & 0.804 & 0.796 & 0.798 & 0.794 \\
% \phantom{1}64\% & 0.817 & 0.652 & 0.817 & 0.648 & 0.731 & 0.725 \\%& 0.804 & 0.798 & 0.809 & 0.793 & 0.806 & 0.797 \\%& 0.799 & 0.791 & 0.814 & 0.791 & 0.814 & 0.799 \\
Empty hyp. & \multicolumn{2}{c}{0.316} & \multicolumn{2}{c}{0.329} & \multicolumn{2}{c}{0.391} \\
Random hyp. & \multicolumn{2}{c}{0.463} & \multicolumn{2}{c}{0.472} & \multicolumn{2}{c}{0.435} \\
 \bottomrule
 \end{tabular}
 % }
 \vspace{-1ex}
 \caption{COMET$^\mathrm{DA}_\mathrm{22}$ scores for WMT23 Online-A's output in (1) correct, (2, 3) incorrect language, (4) empty outputs, (5) random, but fluent, output.}\vspace{-1ex}
 \label{tab:mismatch}
\end{table}

\input{tables/avg-wmt-lang-score.tex}

\subsection{Score distribution bias  \problemtypedata}

As the COMET metric is a machine learning model trained on human ratings of existing machine translations, it inherits many properties of statistical learning such as data (output score) distribution bias. The yearly WMT shared task receives submissions with varying quality and potentially varying quality ranges for different translation directions per year (from \citealp{koehn-monz-2006-manual} to \citealp{freitag-etal-2023-results}). 
This is attributed to diverse factors: the availability of data, source-target language similarity, the level of interest in languages, etc. 
The gap in translation quality will then propagate into skewed human judgement scores across translation directions.
When a single COMET model learns to score all translation directions, it can overfit the score distribution w.r.t. a translation direction in addition to the quality of a translation hypothesis.

\begin{figure}[t]
\centering
\includegraphics[width=0.4\linewidth]{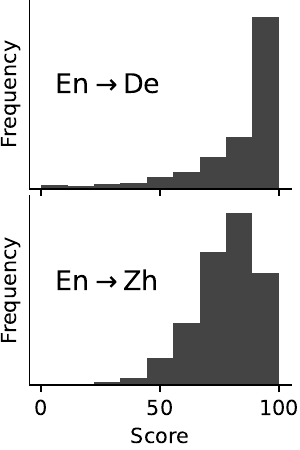}
\raisebox{20mm}{\scalebox{2}{$\rightarrow$}}
\includegraphics[width=0.4\linewidth]{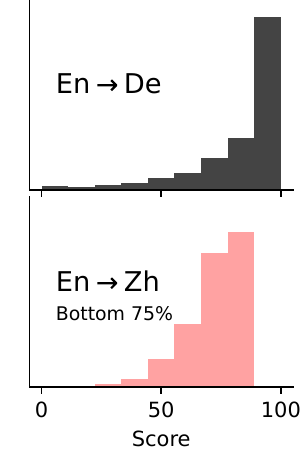}
\vspace{-1ex}
\caption{Setup of an experiment with bottom 75\% of \EnZh{} scores which creates a bias in COMET$_\mathrm{22}^\mathrm{DA}$. In the new data for \EnZh{} (bottom right) there are no translations with perfect scores. \EnDe{} data are unaffected.}\vspace{-1ex}
\label{fig:distribution_bias_setup}
\end{figure}

We first verify this in \Cref{tab:distribution_bias_per_lang} which shows that WMT translation directions are associated with vastly different human DA scores (from 0.51 to 0.91). 
Empirically, we illustrate this issue using two high-resource directions \EnDe{} and En$\rightarrow$Zh. 
As shown in \Cref{fig:distribution_bias_setup}, we keep either the top- or bottom-75\% of all scored translations to alter the score distribution for each direction, simulating the scenario where low- and high-performing system submissions are received for different directions. 
We then train different COMET models on the human train data before and after alteration as per \Cref{fig:distribution_bias_setup}. 
Finally, we evaluate those checkpoints on the same test set and report results in \Cref{tab:distribution_bias_results}. 

As expected, for both \EnDe{} and En$\rightarrow$Zh, training on the top or bottom-scoring data leads to increased or decreased COMET scores on the same set of hypotheses.
Besides, we observe that altering scores in a particular translation direction inconsistently affects scores in another direction systematically.
For example, removing the bottom 25\% \EnDe{} scores ``improves'' the test-time score from 0.770 to 0.790 whilst En$\rightarrow$Zh remains unaffected.

Finally, an empty translation should have the same score irrespective of the original source.
Due to the statistical learning nature of the metric, this is not the case, as found in \Cref{sec:empty_hypothesis}.
\citet{sun-etal-2020-estimating} and \citet{zouhar-etal-2023-poor} show that some meaningful correlation with human scores can be attained with just the source as the input.
This shows, that there is a learned bias based on the prior difficulty of the source segment, which is undesirable for an objective evaluation metric.

\begin{table}[t]
\centering\small
\begin{tabular}{l@{\,}lcccc}
\toprule
\multicolumn{2}{c}{Training data} & \EnDe & \EnZh \\
\midrule
All & & \cellcolor{gray!25} 0.770 & \cellcolor{gray!25} 0.770 \\
Top-75\% & of \EnDe & \cellcolor{blue!25} 0.790 & 0.770 \\
Bot-75\% & of \EnDe & \cellcolor{red!25} 0.765 & 0.764 \\
Top-75\% & of \EnZh & 0.783 & \cellcolor{blue!25} 0.789 \\
Bot-75\% & of \EnZh & 0.772 & \cellcolor{red!25} 0.751 \\
\bottomrule
\end{tabular}\vspace{-1ex}
\caption{Average scores from COMET$_\mathrm{22}^\mathrm{DA}$ trained on data with \hlc[blue!25]{top-} or \hlc[red!25]{bottom-}75\% of scores kept in a particular direction.}
\vspace{-2ex}
\label{tab:distribution_bias_results}
\end{table}

\paragraph{Implication.}
A trivial conclusion is that COMET scores for different translation directions are not comparable. Nevertheless, we caution that the same phenomenon could happen for other features such as the domain, output style, etc.
Although z-score re-scaling \textit{could} mitigate this problem, it has not been a common practice since WMT22 and it would further contribute to non-objective scores \citep{knowles-2021-stability}. Moreover, while z-scoring is straightforward for the translation direction, other latent, language-agnostic biases still exist.

\subsection{Domain bias \problemtypedata}

Neural metrics like COMET are biased towards particular domains, manifested by worse test performance on unseen domains \citep{zouhar2024struggle}.
Taking inspiration from previous work and our discussions on ``latent biases'', we now raise a question---can we create adversarial hypotheses at test time to exploit the domain bias in training time?
Specifically, different domains in the training data are associated with different score ranges.
By pretending that a translation is in a particular domain, it might manipulate its COMET score.

To make it explicit to COMET during training, we prepend the target translation with a tag of its domain---in our case, the year the WMT data originated. 
Note that in each iteration of WMT, systems get higher overall DA scores (e.g. 0.721 for 2019 and 0.749 for 2021). 
\Cref{tab:domainyear_bias} illustrates our setup: during training, we tag the scored translation data with its year; during testing, we trial various year prefixes to understand the effect. 

\begin{table}[t]
\small\centering
\begin{tabular}{ll}
\toprule
\textbf{Train} &
\texttt{2020} Fire prevented from spreading \\[1ex]
\textbf{Test} & 
\texttt{2019} Now I have to tell you a nice story. \\
& \texttt{2020} Now I have to tell you a nice story. \\
& \texttt{2021} Now I have to tell you a nice story. \\
& \texttt{2022} Now I have to tell you a nice story. \\
& \texttt{2023} Now I have to tell you a nice story. \\
\bottomrule
\end{tabular}\vspace{-1ex}
\caption{An illustration of year-as-a-domain tagging during training and testing.}\vspace{-1ex}
\label{tab:domainyear_bias}
\end{table}

\begin{table}[t]
\centering\small\setlength{\tabcolsep}{3ex}
\begin{tabular}{lcc}
\toprule
 Tag & Train & Test \\
\midrule
2018 & unseen & 0.736 \\
2019 & 0.721 & 0.737 \\
2020 & 0.735 & 0.744 \\
2021 & 0.749 & 0.749 \\
2022 & unseen & 0.747 \\
2023 & unseen & 0.747 \\
2024 & unseen & 0.739 \\
2025 & unseen & 0.747 \\
\bottomrule
\end{tabular}\vspace{-1ex}
\caption{Average COMET$_\mathrm{22}^\mathrm{DA}$ scores for subsets in training and predictions on test data. During testing, the whole test set had a single tag, e.g. 2024, irrespective of the data origin.}
\label{tab:domainyear_bias_results}
\vspace{-3mm}
\end{table}

One would expect the metric to produce the same score based solely on the translation quality.
However, as shown in \Cref{tab:domainyear_bias_results}, by merely changing the year tag, we can influence the average score of the test set.
During training, the model would be able to observe that 2019 is associated with the worst score and 2021 the best.
During test time, the model follows this bias and also extrapolates it to upcoming years where it predicts an improvement in the average DA scores.
While the differences appear small, they are on the same scale as the differences between years in the training data.

\paragraph{Implication.} 
Our year-as-a-domain setting might be overly simple, but the vulnerability of COMET to latent biases cannot be neglected. Although \citet{amrhein-sennrich-2022-identifying} has shown that COMET is not sensitive to numbers, this work reveals that it can be systematically exploited in an artificial setting. We offer a more practical (adversarial) example that one may disguise biomedical domain translations as news translations to game COMET.

\subsection{Lack of multi-reference support \problemtypeusage}
\label{sec:multiple_references}
In machine translation, there usually exist many valid translations for the same input. An effective metric should incorporate multiple ground truths if available, thereby enhancing the accuracy and robustness of its evaluation. Existing metrics like BLEU or chrF rely on surface-level overlap to capture the ground truth space from multiple references while metrics like ParBLEU \citep{bawden-etal-2020-parbleu} can automatically generate paraphrases of a given reference to be included during evaluation.

By design, only one reference can be used in COMET.
Whilst one may argue that representing a text reference in the neural space can ease the restriction on word choices, it might still be beneficial to use multiple references to overcome defects in the base embedding model.
Therefore, we test whether COMET can explicitly and reliably leverage multiple references.
We identify three distinct ways in which multiple references have been incorporated in COMET in previous literature \citep{rei-etal-2020-unbabels, zouhar-bojar-2024-quality}:
\begin{itemize}
 \item \textbf{max}: Taking the maximum over the scores from multiple passes with different references.
 \item \textbf{avg}: Averaging the scores from multiple passes with different references.
 \item \textbf{agg}: Obtaining an aggregate score per example as follows. A quadruplet of source $\mathbf{s}$, hypothesis $\mathbf{h}$, reference $\mathbf{r}$, and alternative reference $\mathbf{\hat{r}}$ is fed to COMET six times in different [src,~hyp,~ref] arrangements: $[\mathbf{s}, \mathbf{h}, \mathbf{r}], [\mathbf{r}, \mathbf{h}, \mathbf{s}], [\mathbf{s}, \mathbf{h}, \mathbf{\hat{r}}], [\mathbf{\hat{r}}, \mathbf{h}, \mathbf{s}], [\mathbf{r}, \mathbf{h}, \mathbf{\hat{r}}]$, as well as $[\mathbf{\hat{r}}, \mathbf{h}, \mathbf{r}]$. Then, the average score from these six passes is multiplied by~$(1-\sigma)$~where $\sigma$ denotes the standard deviation. %See \citet{rei-etal-2020-unbabels} for further explanation.
\end{itemize}

\input{tables/multi-ref-main.tex}

\smallskip
We use additional references from the WMT23 test set if available ({He}$\rightarrow${En}) or the outputs from the best-scoring system in each direction in the metrics shared task \citep{freitag-etal-2023-results} as an alternative reference. We report pairwise system-level accuracy \citep{kocmi-etal-2021-ship} for various translation directions in \Cref{tab:multi-ref-main}. Our results suggest that there is no single method that can consistently take advantage of the inclusion of multiple references with the existing COMET implementation. At a higher inference cost, the six-pass aggregation with COMET might have a tiny edge over other methods when MQM is treated as human ground truths, but it is also outperformed under DA by single-reference or other multi-reference methods.

As translation systems have greatly improved lately, the above pattern might be explained by \citet{freitag-etal-2020-bleu}'s finding that high-quality translation outputs do not benefit from multi-reference evaluation. We also caution that these observations are highly dependent on the quality of the underlying references. As studied previously, obtaining high-quality references is not trivial \citep{freitag-etal-2020-bleu,freitag-etal-2023-results,zouhar-bojar-2024-quality}. Our use of the top-performing system outputs as alternate references is fit for the purpose but not optimal.

\paragraph{Recommendation.}
Our recommendations for the inclusion of multiple references into COMET or even other neural metrics are aspirational as this topic warrants further investigation. Extending unified pre-training \citep{wan-etal-2022-unite} with multiple references in the architecture as well as using training objectives more suitable for handling more than one references \citep{zheng-etal-2018-multi,fomicheva-etal-2020-multi} can be helpful.

\subsection{Translationese \problemtypedata}

COMET has been trained with human translations as references and machine translations as hypotheses, where both could be deemed ``translationese'' to a certain extent \citep{translationese}.
% An intriguing question is 
% We test whether COMET would be affected by the degree of translationese.

\paragraph{Translationese in references.}

We first conduct an experiment to see if the translationese present in the reference would undermine system evaluation with COMET.
We consider WMT's official reference as a standard version and \citet{freitag-etal-2020-bleu}'s paraphrased reference as a less translationese reference (we use their ``paraphrased as-much-as-possible'' version).
Experiments are carried out under two settings:
(1) WMT19 \EnDe{} submissions scored by COMET$^\mathrm{DA}_\mathrm{22}$, and
(2) WMT20 \EnDe{} submissions scored by COMET$^\mathrm{DA}_\mathrm{20}$ \citep{rei-etal-2020-unbabels}.
These two settings cover two scenarios---whether the test suite has been used in training the COMET model, or not.
A breakdown of COMET scores and rankings for individual systems are listed in \Cref{app:translationese}~\Cref{tab:translationese-in-ref-wmt20,tab:translationese-in-ref-wmt19}.

The COMET scores decline dramatically when we switch the reference from the original one to the paraphrased one---aiming to reduce translationese.
It means that COMET is indeed sensitive to such changes in the reference.
Yet interestingly, the overall system ranking in either setting remains rather stable.
We find a very high Kendall's $\tau_{a}$ of 0.9827 and 0.9833 on the system rankings in the two settings; pairwise accuracy computed against human judgements also maintained at 0.924.
We conclude that translationese in the reference impacts absolute COMET scores but not system ranking.
These patterns are consistent regardless of whether the model has been exposed to the test set.

\begin{figure}[t]  
\begin{mdframed}[style=mdframedlisting]
\begin{lstlisting}
Please paraphrase the following text as
much as possible. Provide the paraphrase
without any explanation:
  
$HYPOTHESIS
\end{lstlisting}
\vspace{-1ex}
\end{mdframed}
\vspace{-1.5ex}
\caption{Prompt template used to request a paraphrase from GPT-4o, where \texttt{\$HYPOTHESIS} is replaced by individual hypotheses.}
\label{fig:gpt-paraphrase-instruction}

\vspace{-2ex}
\end{figure}

\paragraph{Translationese in hypotheses.}
We then attempt to understand if a varying degree of translationese in the system outputs will influence system ranking by COMET. We run COMET$^\mathrm{DA}_\mathrm{22}$ and COMET$^\mathrm{kiwiDA}_\mathrm{22}$ on WMT19 \EnDe{} system outputs as well as their corresponding rephrased outputs against the same source and reference. To acquire paraphrases affordably, we shortlist the top-10 systems' translations in the previous experiment and we prompt GPT-4o using the prompt outlined in \Cref{fig:gpt-paraphrase-instruction}.\footnote{We accessed \texttt{gpt-4o-2024-08-06} via API in Aug 2024.} We do not feed the source sentence to prevent the model from revising the quality.

% COMET$^\mathrm{DA}_\mathrm{22}$ and COMET$^\mathrm{kiwiDA}_\mathrm{22}$
Appendix~\Cref{tab:translationese-in-hyp-wmt19} shows that both models yield the same system ranking when the original hypotheses are scored.
After substituting the hypotheses with their paraphrases, the ranking has changed more under COMET$^\mathrm{kiwiDA}_\mathrm{22}$ which witnesses much lower pairwise accuracy and Kendall's $\tau_{c}$ compared to COMET$^\mathrm{DA}_\mathrm{22}$. 
This suggests that COMET$^\mathrm{kiwiDA}_\mathrm{22}$ is more sensitive to potential changes in the degree of translationese than COMET$^\mathrm{DA}_\mathrm{22}$.
% , but we are unable to measure accuracy or correlation w.r.t. human judgements due to the lack of such.

\paragraph{Considerations.}
We note the limitations of our experiments.
% the experimental design.
First, we assume that the paraphrased references are as good as the original ones and less translationese, but we did not verify this when paraphrasing the hypotheses.
If the hypothesis quality has been affected, we also assume that the LLM paraphrasing process affects all system outputs in an equal magnitude.
% that does not affect ranking.
Second, the evaluations that anchor to human judgements assume that human evaluators provide assessment solely on the quality and do not overly insist on adequacy/translationese.
Third, our comparison between COMET$^\mathrm{DA}_\mathrm{22}$ and COMET$^\mathrm{kiwiDA}_\mathrm{22}$ only shows that they do not behave the same in dealing with change in the degree of translationese in hypotheses.
% , without informing us which is more satisfactory.
% Investigating the above problems would be costly but interesting.

\subsection{Model reporting \problemtypeusage}
\label{sec:model_reporting}

Different COMET models can yield distinct results.
Therefore it is important to always specify the specific model for sensible score interpretation and comparison. 
In this section, we examine to what extent this holds up in scientific literature.

We automate this bibliometric task with SemanticScholar API \citep{Kinney2023TheSS}.
Starting with 1100 papers that cite one of the COMET papers, we obtain 417 papers from 2021 to 2024 that have an easily accessible PDF version.\footnote{Papers in 2020 did not have to report the specific model as there was only one available at the time. Further, we acknowledge potential bias to only papers with available PDFs.}
We check if any of the \textit{tables} contains the string \texttt{comet}.
Within those papers, we check whether the COMET model information is contained in the PDF using a regular expression.\footnote{Case-insensitive: ``\texttt{comet[ \textbackslash-](da|20|21|22|23)|wmt\\(20|21|22|23)\textbackslash-comet|xcomet\textbackslash-|wmt\textbackslash-da\textbackslash-estimator}''}
% \footnote{At least one of the following: \texttt{xcomet\textbackslash-}; \\
% \texttt{wmt(20|21|22|23)\textbackslash-comet}; \texttt{wmt\textbackslash-da\textbackslash-estimator}; \\
% \texttt{comet[ \textbackslash-](da|20|21|22|23)}.}
After further manual validation, we found that 50 of the examined papers do not report a specific COMET version.
This establishes that \textit{at least 12\% of papers report COMET scores without specific model information}.

In addition, out of the almost 1000 papers running COMET in their evaluation, most only cite the first COMET paper \citep{rei-etal-2020-comet} instead of the paper that describes the specific models that are being used \citep{rei-etal-2020-unbabels,rei-etal-2022-searching,rei-etal-2022-comet,rei-etal-2022-cometkiwi,rei2023scalingcometkiwiunbabelist2023,rei2023insidestorybetterunderstanding,glushkova-etal-2021-uncertainty-aware,wan-etal-2022-unite,alves2024toweropenmultilinguallarge,guerreiro2023xcomettransparentmachinetranslation}.

\paragraph{Recommendation.}
Always report the COMET version, ideally with a link.
Also, cite the affiliated COMET paper as opposed to the first paper \citep{rei-etal-2020-comet}, because different checkpoints have variations in training regimes that might be crucial in analysing the evaluation outcome.

\subsection{Discussions on other issues}

\paragraph{Interpretation of significance.}
Statistical hypothesis test merely shows how likely the difference between the average of two model's scores %on a test set 
on the same test set is caused by random fluctuation. 
\citet{kocmi2024navigating} shows the significance of a difference between two metric scores can be made arbitrarily high and one can force $p{\rightarrow}0$ by using a sizeable test set. 
This tells little about whether this difference is meaningful to a human reader. 
For this reason, we stress the use of \texttt{mt-thresholds} that converts differences in metric scores to how perceivable they are by human annotators.\footnote{\href{https://kocmitom.github.io/MT-Thresholds/}{kocmitom.github.io/MT-Thresholds}}

% \begin{table}[t]
%  \centering\small
%  \setlength{\tabcolsep}{1ex}
%  \begin{tabular}{lcc}
%  \toprule
%  & \multicolumn{2}{c}{COMET$^\mathrm{DA}_\mathrm{22}$} \\
%  \cmidrule(lr){2-3}
%   & text-davinci-003 & ALMA-7B \\
%  \midrule
%  \XtoX{Cs}{En}  & \textbf{86.16} & 85.90 \\
%  \DeEn & \textbf{84.79} & 83.98 \\
%  \XtoX{Ru}{En} & 84.80 & 84.81 \\
%  \ZhEn & \textbf{81.62} & 79.73 \\
%  \XtoX{Is}{En} & 82.13 & \textbf{85.97} \\
%  \midrule
%  {\{Cs,De,Ru,Zh\}$\rightarrow$En} & \textbf{84.34} & 83.61 \\
%  {\{Cs,De,Is,Ru,Zh\}$\rightarrow$En} & 83.90 & \textbf{84.08} \\
 
%  \bottomrule
%  \end{tabular}
%  \caption{Opposite conclusions can be reached by averaging COMET for different translation directions.\todop{reproduce with WMT23 online systems.}}
% \end{table}

\paragraph{Averaging and subtracting COMET scores.}
Research nowadays favours experiments on multiple translation directions, as multilingual translation models and large language models become trendy. Recent papers are more often seen to report a (macro-)average COMET score as an aggregate measure across many directions, usually in Xx$\rightarrow$En, En$\rightarrow$Xx, and All$\rightarrow$All. Whilst indicative, this is not entirely scientific because (1) the score range is inherently distinct for each translation direction and (2) there is no assurance that the scores are on a linear scale. Consequently, an outlying score in a single direction can distort the average, leading to a false claim. Likewise, absolute COMET score differences are not comparable if from different base numbers or directions. We suggest that, when multiple translation directions are of interest, in addition to averaged scores, practitioners can report the number of wins (against another system) as another aggregation of individual scores.

% a custom to report model performance in multiple translation directions.
% Different translation directions have different inherent ranges. The quality/score of 0.8 COMET for \XtoX{Zh}{En} is not the same as that for \XtoX{De}{En}.
% Hence it does not make scientific sense to aggregate COMET scores for different translation directions to indicate the ``overall'' quality.
% \citet{xu2024a} note that ALMA-7B is superior to GPT-3-D.
% We break down the individual and averaged scores.
% It can be observed that, by leaving one translation direction (\XtoX{Is}{En}) out, the averaged COMET results can support opposite findings.

% \paragraph{Recommendation.} Report individual scores for each direction and consider number of wins.

\paragraph{Optimizing to COMET.} 

Owing to COMET's strong correlation with human judgement, recent works investigate the feasibility of using it in translation modelling directly.
% to aid output quality.
These strategies either include COMET in a distillation workflow \cite{finkelstein2024mbr,guttmann2024chasing}, as a data filtering method \cite{peter-etal-2023-theres}, as a decoding method \cite{freitag-etal-2022-high,fernandes-etal-2022-quality,vernikos-popescu-belis-2024-dont} or as a training objective \cite{yan-etal-2023-bleurt}.

Nevertheless, COMET may cease to be a good measure if practitioners over-optimize a system towards it. As \citet{yan-etal-2023-bleurt} demonstrated, a model trained towards COMET can generate ``universal translations'' (hallucinations) preferred by COMET regardless of the source sentence. Recently, using COMET-based MBR decoding has become prevalent in shared tasks.
% e.g. at the IWSLT24 offline translation task by the CMU team.
% Their system has the highest COMET scores but ranks third in terms of human evaluation in \EnDe{} \cite{ahmad-etal-2024-findings}.
For example, Unbabel-Tower70B at WMT24 also used MBR and dominated all automatic metrics but not so much under human evaluation \citep{kocmi2024preliminarywmt24,kocmi2024wmt}. 
MBR decoding could be seen as an automatic way to exploit bias in the scoring method, (currently COMET in most cases), so practitioners need to be aware of its shortcomings, disclose the use of such, and base system building on multiple (less correlated) metrics \citep{jon-etal-2023-cuni}.

A novel issue is using automated metrics in human evaluation \citep{zouhar2024esaai} that collects data for metric training.
This might create a similar effect as translationese in machine translation.
The data could be biased by the particular quality estimator that is assisting annotators in the data collection process.

\paragraph{Sensitivity to sentence segmentation.}
Like most MT metrics, COMET works at the sentence level, but sometimes sentence-segmented input is not available. This is often the case in speech translation (ST) where sentence segmentation is treated as part of the task \cite{ahmad-etal-2024-findings}. To address the problem of mismatching segmentation between the system output and the reference, a common solution in ST is to re-segment the output using a minimum error rate method \citep{Matusov:2005} in order to force-align it with the reference. Forced alignment can introduce segmentation errors resulting in truncated (thus grammatically incorrect) sentences. There is evidence that COMET, as a metric reliant on sentence embeddings, is more sensitive to segmentation errors than string-based metrics, like BLEU, which rely purely on $n$-gram overlaps with no linguistic notion of a sentence \citep{amrhein-haddow-2022-dont}. In a recent comparison of COMET with human ranking, \citet{sperber-etal-2024-evaluating} suggested that COMET-based ranking is robust to segmentation errors but that a ``more thorough study of this issue is needed''.

\paragraph{Other metrics.}
Our work focused on COMET, the current most popular family of MT metrics. 
Nonetheless, our recommendations could apply to other neural metrics, like MetricX-23 \citep{juraska-etal-2023-metricx}, because many issues we outlined are due to their statistical learning nature. 
Even beyond this, metric reporting and score interpretation in practice, e.g. software usage or averaging across multiple directions, can be problematic for string-matching metrics like BLEU or chrF too.

\section{The \texttt{SacreCOMET} Package}

To help alleviate problems in Sections~\ref{sec:software_version} (software version), \ref{sec:compute_precision} (compute precision), and \ref{sec:model_reporting} (model reporting), we release a simple package \texttt{sacreCOMET} with two functionalities. Given a model name, the first functionality attempts to find the appropriate citation including a link to the paper and a BibTeX:

\begin{mdframed}[style=mdframedlisting]
\begin{lstlisting}
$ pip install sacrecomet
$ sacrecomet cite Unbabel/xcomet-xl

https://arxiv.org/abs/2310.10482
@misc{guerreiro2023xcomet,
 title={xCOMET: Transparent Machine Translation Evaluation through Fine-grained Error Detection}, 
 ...
\end{lstlisting}
\vspace{-1ex}
\end{mdframed}

\noindent The second functionality semi-automatically detects the local software versions to generate a signature for better reproducibility.
Both functionalities can also be run in interactive mode.

\begin{mdframed}[style=mdframedlisting]
\begin{lstlisting}
$ sacrecomet --model unite-mup --prec fp32

Python3.11.8|Comet2.2.2|fp32|unite-mup
\end{lstlisting}
\vspace{-1ex}
\end{mdframed}

\bigskip
\section{Future Work on Learned Metrics}
\medskip

\begin{itemize}[itemsep=1.7ex,left=0mm]
%\item \textbf{Unbiased learning from data}:
% Just because in the training data a system for a particular language pair in a particular domain received lower human scores, test-time scores should not be affected by this distribution bias.
% Future metrics should include training data that is uniformly distributed or improve learning algorithms that mitigate this bias. 
% Future metrics should aim to output more uniform scores such that they are applicable to a range of MT systems with less bias.

\item \textbf{Fixing data bias}:
% Our observations suggest that 
Learned metrics are sensitive to the training data distribution.
Future metrics should aim to reduce the bias caused by the data selection process such that they are applicable to a range of MT systems.

\item \textbf{Interpretability across languages}:
Currently, practitioners cannot compare, or pedantically, aggregate scores in different translation directions. It would be useful to unify the scores to a single scale that can be interpreted independent of the language \citep[similar to][]{kocmi2024navigating}, e.g. to indicate X\% of segments are production-ready.

\item \textbf{Confidence-aware metrics:} As seen in works of \citet{glushkova-etal-2021-uncertainty-aware,10.1162/tacl_a_00330}, it is possible to build metrics that output a confidence interval, though its usage in evaluation and modelling remains scarce.

\item \textbf{Inference speed:} Learned metrics are getting better but at the cost of bulky models and increased inference time. These overheads should be taken into account when developing new models, such as the work of \citet{rei-etal-2022-searching}.

\item \textbf{Representations:} Current COMET models are built upon off-the-shelf multilingual encoder models which are likely trained on human-written texts. However, this could bring in a domain mismatch---when translation hypotheses act as the input to metric models, they are not human-written but machine-translated.

\item \textbf{Robustness}: Metrics should have the correct behaviour even in corner cases, be it empty output or incorrect language. Mapping all inputs \citep[][\it inter alia]{amrhein-etal-2022-aces}, including partial or adversarial ones, evaluating the metrics, and coming up with methods to make them more robust would increase the metrics' adoption and trust.

\item \textbf{Built-in QE:} In production, machine translation and quality estimation are commonly two different processes. In many applications, however, a single QE model is used for a single MT model. Quantifying how much is QE adaptation to a particular MT model useful is beneficial for a holistic understanding of QE metrics. Further, proposing methods for supervised quality estimation built into the MT could ease industry adoption. Beyond the work of \citet{tomani-etal-2024-quality}, this remains largely unexplored.

\item \textbf{Noise-aware training:}
Human annotations are notoriously noisy.
At the scale of WMT data, poor-quality annotations are unavoidable. 
The inter-annotator agreement for even robust annotations, such as ESA, remains low at $\tau_c{\approx}0.3$. 
The effect of data quality on learned metrics is so far unknown and methods for noise/uncertainty-aware training are under-studied.
\end{itemize}

\section{Conclusion}

\vspace{-1mm}

COMET is currently one of the most powerful automatic metrics/quality estimators for machine translation, consistently achieving the top correlation with human judgement. 
In comparison to previous metrics, it is a statistical learning model and thus inherits all the related problems in addition to possible technical misconfigurations. 
We urge practitioners to consider more deeply the use of COMET in non-standard scenarios especially where such training bias might come into play. 
Beyond these issues, there has been confusion in the literature in reporting the correct COMET model and its correct setting. 
For improved consistency, we release an easy-to-use tool to assist practitioners.

\section*{Acknowledgments}
The work has received funding from UK Research and Innovation (UKRI) under the UK government’s Horizon Europe funding guarantee [grant numbers 10052546 and 10039436]. 
We thank the anonymous reviewers for their insightful suggestions.

\bibliography{custom}

\clearpage

\appendix

\section{Batch size and GPU/CPU}
\label{sec:gpucpu_batch_size}

We run the test inference on a combination of GPU or CPU with varying batch sizes (BS, 1 or 100).
Results in \Cref{tab:gpucpu_batch_size} demonstrate that the tiny effects of these choices are negligible for COMET reporting.

\begin{table}[htbp]
\centering\small
\begin{tabular}{l@{\,}c@{\,}lc}
\toprule
\multicolumn{3}{c}{\bf Difference} & \bf MAE \\
\midrule
BS=1, GPU  &$-$& BS=1, GPU  & $0$ \\
BS=1, GPU  &$-$& BS=64 GPU & $2\times10^{-7}$ \\
BS=1, GPU  &$-$& BS=1, CPU  & $4\times10^{-7}$ \\
BS=64, GPU &$-$& BS=64, CPU & $4\times10^{-7}$ \\
\bottomrule
\end{tabular}
\caption{
MAE between segment-level COMET$^\mathrm{DA}_\mathrm{22}$ scores under various inference settings.
The ``{BS=1, GPU}'' setting in the first row was run twice.}
\label{tab:gpucpu_batch_size}
\end{table}

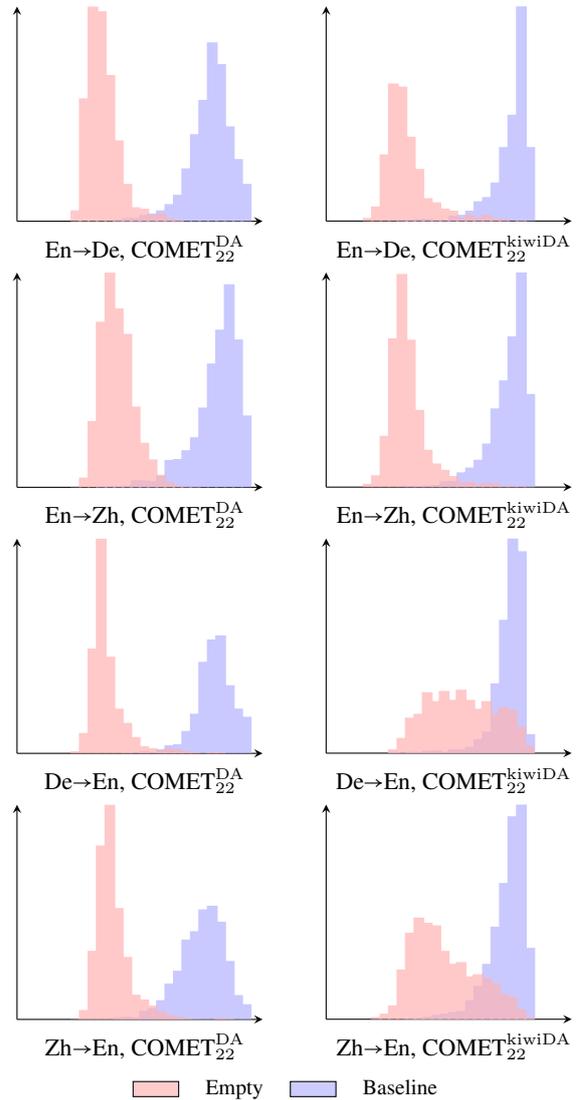
\begin{figure}[t]
\section{Distribution of COMET scores for empty and valid hypothesis}\label{app:empty-distribution}

\centering\small
\input{img/plot-en-de-wmt22da.tex}\hfill%
\input{img/plot-en-de-wmt22da-kiwi.tex}\\
\input{img/plot-en-zh-wmt22da.tex}\hfill%
\input{img/plot-en-zh-wmt22da-kiwi.tex}\\
\input{img/plot-de-en-wmt22da.tex}\hfill%
\input{img/plot-de-en-wmt22da-kiwi.tex}\\
\input{img/plot-zh-en-wmt22da.tex}\hfill%
\input{img/plot-zh-en-wmt22da-kiwi.tex}\\
\input{img/distributions_custom_legend.tex}
\caption{Distribution of instance-level scores for empty and baseline translations (x-axis: score; y-axis: count).}
\label{fig:empty-distribution}
\vspace{7.5cm}
\end{figure}

\clearpage

\begin{table*}[htb]
\section{System rankings before and after paraphrasing references or hypotheses}
\label{app:translationese}
\bigskip

\centering\small
\begin{tabular}{lcccc}
\toprule
\multicolumn{1}{c}{\multirow[b]{2}{*}{WMT20 \EnDe{}}} & \multicolumn{2}{c}{COMET$^\mathrm{DA}_\mathrm{20}$} & \multicolumn{2}{c}{Ranking} \\
\cmidrule(lr){2-3}\cmidrule(lr){4-5}
 & original ref & paraphrased ref & original ref & paraphrased ref \\
\midrule
Sys-1069 & 0.508 & 0.305  & \phantom{0}9  & \phantom{0}9  \\
Sys-832 & 0.540 & 0.333  & \phantom{0}8  & \phantom{0}8  \\
Sys-1535 & 0.560 & 0.356  & \phantom{0}5  & \phantom{0}5  \\
Online-A & 0.499 & 0.288  & 10 & 10 \\
Online-B & 0.554 & 0.351  & \phantom{0}\textbf{7}  & \phantom{0}\textbf{6}  \\
Online-G & 0.268 & 0.052  & 14 & 14 \\
Online-Z & 0.329 & 0.121  & 13 & 13 \\
Sys-73 & 0.405 & 0.192  & 12 & 12 \\
Sys-1520 & 0.563 & 0.360  & \phantom{0}4  & \phantom{0}4  \\
Sys-890 & 0.578 & 0.371  & \phantom{0}3  & \phantom{0}3  \\
Sys-1136 & 0.472  & 0.264  & 11 & 11 \\
Sys-388 & 0.102 & -0.074\phantom{-} & 16 & 16 \\
Sys-737 & 0.555 & 0.351  & \phantom{0}\textbf{6}  & \phantom{0}\textbf{7}  \\
Ref-A & 0.878  & 0.526  & \phantom{0}1  & \phantom{0}1  \\
Ref-B & 0.591 & 0.446  & \phantom{0}2  & \phantom{0}2  \\
Sys-179 & 0.189 & 0.000    & 15 & 15 \\
\cmidrule(lr){4-5}
& & & \multicolumn{2}{c}{$\tau_{c}=0.9833$} \\
& & & $\text{Acc}=0.924$ & $\text{Acc}=0.924$ \\
\bottomrule
\end{tabular}
\caption{Results for WMT20 \EnDe{} submissions evaluated against the original or human-paraphrased reference. Kendall's $\tau_{c}$ is measured between two evaluations based on original and paraphrased references; pairwise system-level accuracy (Acc) is measured against human DA scores.}
\label{tab:translationese-in-ref-wmt20}
\end{table*}

\begin{table*}[ht]
\centering\small
\begin{tabular}{lcccc}
\toprule
\multicolumn{1}{c}{\multirow[b]{2}{*}{WMT19 \EnDe{}}} & \multicolumn{2}{c}{COMET$^\mathrm{DA}_\mathrm{22}$} & \multicolumn{2}{c}{Ranking} \\
\cmidrule(lr){2-3}\cmidrule(lr){4-5}
 & original ref & paraphrased ref & original ref & paraphrased ref \\
\midrule
Sys-6862 & 0.867 & 0.817 & \phantom{0}2 & \phantom{0}2 \\
Sys-6820 & 0.834 & 0.780 & 12 & 12 \\
Sys-6819 & 0.843 & 0.789 & \phantom{0}9 & \phantom{0}9 \\
Sys-6651 & 0.847 & 0.794 & \phantom{0}7 & \phantom{0}7 \\
Sys-6926 & 0.852 & 0.797 & \phantom{0}\textbf{4} & \phantom{0}\textbf{5} \\
Sys-6808 & 0.869 & 0.818 & \phantom{0}1 & \phantom{0}1 \\
Sys-6785 & 0.837 & 0.785 & 11 & 11 \\
Sys-6974 & 0.866 & 0.814 & \phantom{0}3 & \phantom{0}3 \\
Sys-6763 & 0.851 & 0.797 & \phantom{0}\textbf{5} & \phantom{0}\textbf{6} \\
Sys-6674 & 0.811 & 0.756 & 17 & 17 \\
Sys-6508 & 0.804 & 0.752 & 19 & 19 \\
Sys-6731 & 0.850 & 0.797 & \phantom{0}\textbf{6} & \phantom{0}\textbf{4} \\
Sys-6871 & 0.809 & 0.756 & 18 & 18 \\
Sys-6479 & 0.833 & 0.777 & 13 & 13 \\
Sys-6823 & 0.845 & 0.792 & \phantom{0}8 & \phantom{0}8 \\
Sys-6790 & 0.386 & 0.364 & 22 & 22 \\
Sys-6981 & 0.826 & 0.774 & 14 & 14 \\
Online-A & 0.815 & 0.761 & 16 & 16 \\
Online-B & 0.838 & 0.784 & 10 & 10 \\
Online-G & 0.795 & 0.738 & 20 & 20 \\
Online-X & 0.728 & 0.673 & 21 & 21 \\
Online-Y & 0.821 & 0.762 & 15 & 15 \\
\cmidrule(lr){4-5}
& & & \multicolumn{2}{c}{$\tau_{c}=0.9827$} \\
& & & $\text{Acc}=0.875$ & $\text{Acc}=0.845$ \\
\bottomrule
\end{tabular}
\caption{Results for WMT19 \EnDe{} submissions evaluated against the original or human-paraphrased reference. Kendall's $\tau_{c}$ is measured between two evaluations based on original and paraphrased references; pairwise system-level accuracy (Acc) is measured against human DA scores.}

\label{tab:translationese-in-ref-wmt19}
\end{table*}

\begin{table*}[t]
\centering\small
\begin{tabular}{lcccccccc}
\toprule
\multicolumn{1}{c}{\multirow[b]{2}{*}{\makecell{WMT19\\\EnDe}}} & \multicolumn{2}{c}{COMET$^\mathrm{DA}_\mathrm{22}$} & \multicolumn{2}{c}{Ranking} & \multicolumn{2}{c}{COMET$^\mathrm{kiwiDA}_\mathrm{22}$} & \multicolumn{2}{c}{Ranking} \\
\cmidrule(lr){2-5}\cmidrule(lr){6-9}
 & {orig. hyp} & {para. hyp} & {orig. hyp} & {para. hyp} & {orig. hyp} & {para. hyp} & {orig. hyp} & {para. hyp} \\
\midrule
Sys-6823 & 0.845 & 0.840 & \phantom{1}\textbf{8} & \phantom{1}\textbf{5} & 0.821 & 0.824 & \phantom{1}\textbf{8} & \phantom{1}\textbf{4} \\
Sys-6862 & 0.867 & 0.842 & \phantom{1}2 & \phantom{1}2 & 0.840 & 0.828 & \phantom{1}\textbf{2} & \phantom{1}\textbf{3} \\
Sys-6819 & 0.843 & 0.835 & \phantom{1}9 & \phantom{1}9 & 0.815 & 0.817 & \phantom{1}\textbf{9} & \phantom{1}\textbf{8} \\
Sys-6808 & 0.869 & 0.843 & \phantom{1}1 & \phantom{1}1 & 0.840 & 0.828 & \phantom{1}\textbf{1} & \phantom{1}\textbf{2} \\
Sys-6974 & 0.866 & 0.842 & \phantom{1}3 & \phantom{1}3 & 0.838 & 0.829 & \phantom{1}\textbf{3} & \phantom{1}\textbf{1} \\
Sys-6651 & 0.847 & 0.835 & \phantom{1}\textbf{7} & \phantom{1}\textbf{8} & 0.821 & 0.817 & \phantom{1}\textbf{7} & \phantom{1}\textbf{9} \\
Sys-6926 & 0.852 & 0.839 & \phantom{1}\textbf{4} & \phantom{1}\textbf{6} & 0.824 & 0.822 & \phantom{1}\textbf{4} & \phantom{1}\textbf{6} \\
Sys-6763 & 0.851 & 0.841 & \phantom{1}\textbf{5} & \phantom{1}\textbf{4} & 0.823 & 0.824 & \phantom{1}5 & \phantom{1}5 \\
Online-B & 0.838 & 0.830 & 10 & 10 & 0.805 & 0.810 & 10 & 10 \\
Sys-6731 & 0.850 & 0.837 & \phantom{1}\textbf{6} & \phantom{1}\textbf{7} & 0.822 & 0.820 & \phantom{1}\textbf{6} & \phantom{1}\textbf{7} \\
\cmidrule(lr){4-5}\cmidrule(lr){8-9}
& & & \multicolumn{2}{r}{$\tau_{c}=0.822$} & & & \multicolumn{2}{r}{$\tau_{c}=0.644$} \\
& & & \multicolumn{2}{r}{$\text{Acc}=0.911$} & & & \multicolumn{2}{r}{$\text{Acc}=0.822$} \\
\bottomrule
\end{tabular}
\caption{Results for WMT19 \EnDe{} system outputs and LLM-paraphrased outputs evaluated against the original reference. Both Kendall's $\tau_{c}$ and pairwise system-level accuracy (Acc) are measured between two evaluations based on original and paraphrased references.}
\label{tab:translationese-in-hyp-wmt19}
\end{table*}

\end{document}

%% file: img/plot-zh-en-wmt22da.tex
 % ZhEn, COMET$^\mathrm{DA}_\mathrm{22}$
 % Automatically generated. Do not modify.
\begin{tikzpicture}
\begin{axis}[
    ybar,
    bar width=1ex,
    width=0.625\linewidth,
    height=0.575\linewidth,
    axis x line=bottom,
    axis y line=left,
    xlabel={\ZhEn, COMET$^\mathrm{DA}_\mathrm{22}$},
    xmin=0,
    xmax=100,
    xtick={},
    xticklabels={},
    xtick style={draw=none},
    ymin=0,
    ytick={},
    yticklabels={},
    ytick style={draw=none},
    label style={font=\footnotesize, xshift=0.5ex, yshift=1.5ex},
]
\addplot+[opacity=0.7,draw=none] coordinates {
(44.8, 6.0)
(48.3, 7.0)
(51.7, 5.0)
(55.2, 24.0)
(58.6, 31.0)
(62.1, 53.0)
(65.5, 99.0)
(69.0, 134.0)
(72.4, 220.0)
(75.9, 243.0)
(79.3, 303.0)
(82.8, 313.0)
(86.2, 269.0)
(89.7, 161.0)
(93.1, 67.0)
(96.6, 42.0)
};
\addplot+[opacity=0.7,draw=none] coordinates {
(20.7, 1.0)
(24.1, 26.0)
(27.6, 189.0)
(31.0, 479.0)
(34.5, 591.0)
(37.9, 307.0)
(41.4, 172.0)
(44.8, 68.0)
(48.3, 55.0)
(51.7, 38.0)
(55.2, 22.0)
(58.6, 12.0)
(62.1, 8.0)
(65.5, 2.0)
(69.0, 1.0)
(79.3, 4.0)
(82.8, 2.0)
};
% \legend{baseline, empty}
\end{axis}
\end{tikzpicture}

%% file: img/plot-zh-en-wmt22da-kiwi.tex
 % ZhEn, COMET$^\mathrm{kiwiDA}_\mathrm{22}$
 % Automatically generated. Do not modify.
\begin{tikzpicture}
\begin{axis}[
    ybar,
    bar width=1ex,
    width=0.625\linewidth,
    height=0.575\linewidth,
    axis x line=bottom,
    axis y line=left,
    xlabel={\ZhEn, COMET$^\mathrm{kiwiDA}_\mathrm{22}$},
    xmin=0,
    xmax=100,
    xtick={},
    xticklabels={},
    xtick style={draw=none},
    ymin=0,
    ytick={},
    yticklabels={},
    ytick style={draw=none},
    label style={font=\footnotesize, xshift=0.5ex, yshift=1.5ex},
]
\addplot+[opacity=0.7,draw=none] coordinates {
(37.9, 1.0)
(41.4, 3.0)
(44.8, 6.0)
(48.3, 10.0)
(51.7, 13.0)
(55.2, 17.0)
(58.6, 19.0)
(62.1, 44.0)
(65.5, 77.0)
(69.0, 128.0)
(72.4, 214.0)
(75.9, 315.0)
(79.3, 469.0)
(82.8, 496.0)
(86.2, 165.0)
};
\addplot+[opacity=0.7,draw=none] coordinates {
(17.2, 5.0)
(20.7, 13.0)
(24.1, 32.0)
(27.6, 115.0)
(31.0, 200.0)
(34.5, 235.0)
(37.9, 223.0)
(41.4, 217.0)
(44.8, 158.0)
(48.3, 128.0)
(51.7, 132.0)
(55.2, 99.0)
(58.6, 104.0)
(62.1, 95.0)
(65.5, 83.0)
(69.0, 59.0)
(72.4, 51.0)
(75.9, 23.0)
(79.3, 5.0)
};
% \legend{baseline, empty}
\end{axis}
\end{tikzpicture}

%% file: img/distributions_custom_legend.tex
\makeatletter
\newenvironment{customlegend}[1][]{%
    \begingroup
    \pgfplots@init@cleared@structures
    \pgfplotsset{#1}%
}{%
    \pgfplots@createlegend
    \endgroup
}%
\def\addlegendimage{\pgfplots@addlegendimage}
\makeatother

\begin{tikzpicture}
\begin{customlegend}[
        legend columns=-1,
        legend style={column sep=2ex,
        draw=none,
        nodes={scale=0.9, transform shape}},
        legend entries={Empty,Baseline}
]
\addlegendimage{
        area legend,
        draw=none,
        fill=red!20,
        opacity=30}
\addlegendimage{
        area legend,
        draw=none,
        fill=blue!20,
        opacity=30}
\end{customlegend}
\end{tikzpicture}

%% file: tables/avg-wmt-lang-score.tex
\begin{table}[htbp]
% \section{Average COMET scores for WMT translation directions}\label{app:score-distribution}
\centering\small
% \resizebox{0.32\linewidth}{!}{
\begin{tabular}{p{0.8cm}p{0.8cm}}
\toprule
Lang & Score \\
\midrule
\XtoX{ De }{ En } & 0.754 \\
\XtoX{ Ps }{ En } & 0.670 \\
\XtoX{ Is }{ En } & 0.724 \\
\XtoX{ Pl }{ En } & 0.761 \\
\XtoX{ Ru }{ En } & 0.771 \\
\XtoX{ Ja }{ En } & 0.663 \\
\XtoX{ Ta }{ En } & 0.655 \\
\XtoX{ Zh }{ En } & 0.743 \\
\XtoX{ Ha }{ En } & 0.641 \\
\XtoX{ Km }{ En } & 0.659 \\
\XtoX{ Lt }{ En } & 0.726 \\
\XtoX{ Cs }{ En } & 0.740 \\
\XtoX{ Gu }{ En } & 0.575 \\
\XtoX{ Kk }{ En } & 0.649 \\
\XtoX{ Iu }{ En } & 0.724 \\
\XtoX{ Fi }{ En } & 0.719 \\
\bottomrule
\end{tabular}
% }
% \resizebox{0.32\linewidth}{!}{
\begin{tabular}{p{0.8cm}p{0.8cm}}
\toprule
Lang & Score \\
\midrule
\XtoX{ En }{ Is } & 0.666 \\
\XtoX{ En }{ Lt } & 0.600 \\
\XtoX{ En }{ Ru } & 0.765 \\
\XtoX{ En }{ Iu } & 0.720 \\
\XtoX{ En }{ Ha } & 0.768 \\
\XtoX{ En }{ Ja } & 0.745 \\
\XtoX{ En }{ Pl } & 0.706 \\
\XtoX{ En }{ Gu } & 0.514 \\
\XtoX{ En }{ Cs } & 0.767 \\
\XtoX{ En }{ Zh } & 0.775 \\
\XtoX{ En }{ Fi } & 0.616 \\
\XtoX{ En }{ Ta } & 0.709 \\
\XtoX{ En }{ De } & 0.841 \\
\XtoX{ En }{ Kk } & 0.574 \\
\bottomrule
% padding to go up
\vspace{0mm}\null
\end{tabular}
% }
% \resizebox{0.32\linewidth}{!}{
\begin{tabular}{p{0.8cm}p{0.8cm}}
\toprule
Lang & Score \\
\midrule
\XtoX{ Hi }{ Bn } & 0.910 \\
\XtoX{ De }{ Fr } & 0.792 \\
\XtoX{ Fr }{ De } & 0.834 \\
\XtoX{ Zu }{ Xh } & 0.639 \\
\XtoX{ De }{ Cs } & 0.510 \\
\XtoX{ Xh }{ Zu } & 0.574 \\
\XtoX{ Bn }{ Hi } & 0.770 \\
\bottomrule
% padding to go up
\vspace{24.6mm}\null
\end{tabular}
% }
% COMET$^\mathrm{DA}_\mathrm{22}$ 
\caption{Average human DA score for each translation direction in WMT data up to 2023 (inclusive).}
\label{tab:distribution_bias_per_lang}
\end{table}

%% file: tables/multi-ref-main.tex
% VZ: intentionally [htbp] so that the list above is not weirdly split

\begin{table}[t]
\centering\small
\setlength{\tabcolsep}{1ex}
\begin{tabular}{lcccccc}
\toprule
\multicolumn{2}{c}{\multirow[b]{2}{*}{WMT23}} & \multicolumn{5}{c}{COMET$^\mathrm{DA}_\mathrm{22}$} \\ \cmidrule(lr){3-7} 

 &  & \multicolumn{1}{c}{ref} & \multicolumn{1}{c}{ref (alt)} & \multicolumn{1}{c}{avg} & \multicolumn{1}{c}{max} & \multicolumn{1}{c}{agg} \\
 \midrule
\multirow{3}{*}{MQM} & \XtoX{He}{En} & 0.885          & 0.897          & 0.910          & 0.910          & \textbf{0.949} \\
                     & \EnDe & 0.974          & 0.936          & 0.974          & \textbf{0.987} & 0.974          \\
                     & \ZhEn & 0.783          & 0.908          & 0.850          & 0.858          & \textbf{0.950}  \\
                \midrule
\multirow{8}{*}{DA}  & \EnDe & 0.885          & \textbf{0.949} & 0.910          & 0.897          & 0.910          \\
                     & \ZhEn & 0.717          & 0.875 & 0.783          & 0.775          & \textbf{0.883}          \\
                     & \DeEn & 0.901          & 0.912          & \textbf{0.923} & \textbf{0.923} & 0.912          \\
                     & \EnZh & \textbf{0.933} & 0.817          & 0.900          & 0.867          & \textbf{0.933}          \\
& \XtoX{Cs}{Uk} & 0.846          & 0.802          & \textbf{0.901} & 0.890          & 0.868 \\
& \XtoX{En}{Cs} & 0.858          & \textbf{0.875} & 0.858          & 0.858          & 0.867 \\
& \XtoX{En}{Ja} & \textbf{0.941} & 0.824          & 0.934          & 0.934          & 0.926 \\
& \XtoX{Ja}{En} & 0.922          & 0.915 & \textbf{0.928} & \textbf{0.928} & 0.922 \\
\bottomrule
\end{tabular}\vspace{-1ex}
\caption{Pairwise system-level accuracy for different strategies incorporating multiple references into COMET$^\mathrm{DA}_\mathrm{22}$. Evaluation is carried out on WMT23 with DA or MQM scores as human ground truths.}\vspace{-1ex}
\label{tab:multi-ref-main}
\vspace{-2mm}
\end{table}

%% file: img/plot-en-de-wmt22da.tex
 % EnDe, COMET$^\mathrm{DA}_\mathrm{22}$
 % Automatically generated. Do not modify.
\begin{tikzpicture}
\begin{axis}[
    ybar,
    bar width=1ex,
    width=0.625\linewidth,
    height=0.575\linewidth,
    axis x line=bottom,
    axis y line=left,
    xlabel={\EnDe, COMET$^\mathrm{DA}_\mathrm{22}$},
    xmin=0,
    xmax=100,
    xtick={},
    xticklabels={},
    xtick style={draw=none},
    ymin=0,
    ytick={},
    yticklabels={},
    ytick style={draw=none},
    label style={font=\footnotesize, xshift=0.5ex, yshift=1.5ex},
]
\addplot+[opacity=0.7,draw=none] coordinates {
(44.8, 1.0)
(48.3, 2.0)
(51.7, 2.0)
(55.2, 4.0)
(58.6, 5.0)
(62.1, 7.0)
(65.5, 11.0)
(69.0, 17.0)
(72.4, 34.0)
(75.9, 56.0)
(79.3, 78.0)
(82.8, 115.0)
(86.2, 101.0)
(89.7, 61.0)
(93.1, 40.0)
(96.6, 24.0)
};
\addplot+[opacity=0.7,draw=none] coordinates {
(20.7, 7.0)
(24.1, 79.0)
(27.6, 138.0)
(31.0, 135.0)
(34.5, 94.0)
(37.9, 52.0)
(41.4, 24.0)
(44.8, 9.0)
(48.3, 8.0)
(51.7, 4.0)
(55.2, 5.0)
(58.6, 2.0)
(62.1, 1.0)
};
% \legend{baseline, empty}
\end{axis}
\end{tikzpicture}

%% file: img/plot-en-de-wmt22da-kiwi.tex
 % EnDe, COMET$^\mathrm{kiwiDA}_\mathrm{22}$
 % Automatically generated. Do not modify.
\begin{tikzpicture}
\begin{axis}[
    ybar,
    bar width=1ex,
    width=0.625\linewidth,
    height=0.575\linewidth,
    axis x line=bottom,
    axis y line=left,
    xlabel={\EnDe, COMET$^\mathrm{kiwiDA}_\mathrm{22}$},
    xmin=0,
    xmax=100,
    xtick={},
    xticklabels={},
    xtick style={draw=none},
    ymin=0,
    ytick={},
    yticklabels={},
    ytick style={draw=none},
    label style={font=\footnotesize, xshift=0.5ex, yshift=1.5ex},
]
\addplot+[opacity=0.7,draw=none] coordinates {
(34.5, 1.0)
(37.9, 1.0)
(41.4, 2.0)
(44.8, 2.0)
(48.3, 1.0)
(51.7, 1.0)
(55.2, 5.0)
(58.6, 6.0)
(62.1, 10.0)
(65.5, 19.0)
(69.0, 23.0)
(72.4, 37.0)
(75.9, 71.0)
(79.3, 102.0)
(82.8, 206.0)
(86.2, 71.0)
};
\addplot+[opacity=0.7,draw=none] coordinates {
(13.8, 3.0)
(17.2, 14.0)
(20.7, 64.0)
(24.1, 132.0)
(27.6, 129.0)
(31.0, 81.0)
(34.5, 50.0)
(37.9, 22.0)
(41.4, 18.0)
(44.8, 12.0)
(48.3, 10.0)
(51.7, 3.0)
(55.2, 6.0)
(58.6, 3.0)
(62.1, 6.0)
(65.5, 2.0)
(69.0, 2.0)
(75.9, 1.0)
};
% \legend{baseline, empty}
\end{axis}
\end{tikzpicture}

%% file: img/plot-en-zh-wmt22da.tex
 % EnZh, COMET$^\mathrm{DA}_\mathrm{22}$
 % Automatically generated. Do not modify.
\begin{tikzpicture}
\begin{axis}[
    ybar,
    bar width=1ex,
    width=0.625\linewidth,
    height=0.575\linewidth,
    axis x line=bottom,
    axis y line=left,
    xlabel={\EnZh, COMET$^\mathrm{DA}_\mathrm{22}$},
    xmin=0,
    xmax=100,
    xtick={},
    xticklabels={},
    xtick style={draw=none},
    ymin=0,
    ytick={},
    yticklabels={},
    ytick style={draw=none},
    label style={font=\footnotesize, xshift=0.5ex, yshift=1.5ex},
]
\addplot+[opacity=0.7,draw=none] coordinates {
(37.9, 1.0)
(41.4, 1.0)
(44.8, 2.0)
(48.3, 4.0)
(51.7, 15.0)
(55.2, 15.0)
(58.6, 13.0)
(62.1, 25.0)
(65.5, 59.0)
(69.0, 61.0)
(72.4, 80.0)
(75.9, 108.0)
(79.3, 172.0)
(82.8, 267.0)
(86.2, 340.0)
(89.7, 438.0)
(93.1, 319.0)
(96.6, 155.0)
};
\addplot+[opacity=0.7,draw=none] coordinates {
(20.7, 1.0)
(24.1, 21.0)
(27.6, 143.0)
(31.0, 357.0)
(34.5, 463.0)
(37.9, 386.0)
(41.4, 331.0)
(44.8, 174.0)
(48.3, 96.0)
(51.7, 60.0)
(55.2, 27.0)
(58.6, 11.0)
(62.1, 3.0)
(65.5, 1.0)
(69.0, 1.0)
};
% \legend{baseline, empty}
\end{axis}
\end{tikzpicture}

%% file: img/plot-en-zh-wmt22da-kiwi.tex
 % EnZh, COMET$^\mathrm{kiwiDA}_\mathrm{22}$
 % Automatically generated. Do not modify.
\begin{tikzpicture}
\begin{axis}[
    ybar,
    bar width=1ex,
    width=0.625\linewidth,
    height=0.575\linewidth,
    axis x line=bottom,
    axis y line=left,
    xlabel={\EnZh, COMET$^\mathrm{kiwiDA}_\mathrm{22}$},
    xmin=0,
    xmax=100,
    xtick={},
    xticklabels={},
    xtick style={draw=none},
    ymin=0,
    ytick={},
    yticklabels={},
    ytick style={draw=none},
    label style={font=\footnotesize, xshift=0.5ex, yshift=1.5ex},
]
\addplot+[opacity=0.7,draw=none] coordinates {
(37.9, 1.0)
(41.4, 3.0)
(44.8, 4.0)
(48.3, 6.0)
(51.7, 16.0)
(55.2, 17.0)
(58.6, 40.0)
(62.1, 53.0)
(65.5, 74.0)
(69.0, 118.0)
(72.4, 213.0)
(75.9, 289.0)
(79.3, 408.0)
(82.8, 581.0)
(86.2, 252.0)
};
\addplot+[opacity=0.7,draw=none] coordinates {
(13.8, 9.0)
(17.2, 32.0)
(20.7, 139.0)
(24.1, 403.0)
(27.6, 577.0)
(31.0, 399.0)
(34.5, 209.0)
(37.9, 101.0)
(41.4, 64.0)
(44.8, 47.0)
(48.3, 30.0)
(51.7, 14.0)
(55.2, 14.0)
(58.6, 7.0)
(62.1, 11.0)
(65.5, 10.0)
(69.0, 3.0)
(72.4, 3.0)
(75.9, 2.0)
(79.3, 1.0)
};
% \legend{baseline, empty}
\end{axis}
\end{tikzpicture}

%% file: img/plot-de-en-wmt22da.tex
 % DeEn, COMET$^\mathrm{DA}_\mathrm{22}$
 % Automatically generated. Do not modify.
\begin{tikzpicture}
\begin{axis}[
    ybar,
    bar width=1ex,
    width=0.625\linewidth,
    height=0.575\linewidth,
    axis x line=bottom,
    axis y line=left,
    xlabel={\DeEn, COMET$^\mathrm{DA}_\mathrm{22}$},
    xmin=0,
    xmax=100,
    xtick={},
    xticklabels={},
    xtick style={draw=none},
    ymin=0,
    ytick={},
    yticklabels={},
    ytick style={draw=none},
    label style={font=\footnotesize, xshift=0.5ex, yshift=1.5ex},
]
\addplot+[opacity=0.7,draw=none] coordinates {
(48.3, 1.0)
(55.2, 2.0)
(62.1, 4.0)
(65.5, 8.0)
(69.0, 9.0)
(72.4, 22.0)
(75.9, 39.0)
(79.3, 90.0)
(82.8, 112.0)
(86.2, 116.0)
(89.7, 67.0)
(93.1, 44.0)
(96.6, 36.0)
};
\addplot+[opacity=0.7,draw=none] coordinates {
(20.7, 2.0)
(24.1, 20.0)
(27.6, 103.0)
(31.0, 211.0)
(34.5, 95.0)
(37.9, 44.0)
(41.4, 30.0)
(44.8, 16.0)
(48.3, 7.0)
(51.7, 5.0)
(55.2, 3.0)
(58.6, 5.0)
(62.1, 4.0)
(65.5, 2.0)
(69.0, 1.0)
(75.9, 1.0)
(79.3, 1.0)
};
% \legend{baseline, empty}
\end{axis}
\end{tikzpicture}

%% file: img/plot-de-en-wmt22da-kiwi.tex
 % DeEn, COMET$^\mathrm{kiwiDA}_\mathrm{22}$
 % Automatically generated. Do not modify.
\begin{tikzpicture}
\begin{axis}[
    ybar,
    bar width=1ex,
    width=0.625\linewidth,
    height=0.575\linewidth,
    axis x line=bottom,
    axis y line=left,
    xlabel={\DeEn, COMET$^\mathrm{kiwiDA}_\mathrm{22}$},
    xmin=0,
    xmax=100,
    xtick={},
    xticklabels={},
    xtick style={draw=none},
    ymin=0,
    ytick={},
    yticklabels={},
    ytick style={draw=none},
    label style={font=\footnotesize, xshift=0.5ex, yshift=1.5ex},
]
\addplot+[opacity=0.7,draw=none] coordinates {
(31.0, 1.0)
(37.9, 1.0)
(41.4, 1.0)
(44.8, 2.0)
(48.3, 2.0)
(51.7, 1.0)
(55.2, 3.0)
(58.6, 3.0)
(62.1, 5.0)
(65.5, 11.0)
(69.0, 17.0)
(72.4, 55.0)
(75.9, 105.0)
(79.3, 169.0)
(82.8, 159.0)
(86.2, 15.0)
};
\addplot+[opacity=0.7,draw=none] coordinates {
(24.1, 3.0)
(27.6, 12.0)
(31.0, 28.0)
(34.5, 31.0)
(37.9, 48.0)
(41.4, 41.0)
(44.8, 49.0)
(48.3, 42.0)
(51.7, 50.0)
(55.2, 42.0)
(58.6, 35.0)
(62.1, 41.0)
(65.5, 31.0)
(69.0, 36.0)
(72.4, 34.0)
(75.9, 21.0)
(79.3, 6.0)
};
% \legend{baseline, empty}
\end{axis}
\end{tikzpicture}